\documentclass[journal,12pt,onecolumn,draftclsnofoot,]{IEEEtran}

\usepackage{cite}
\usepackage{amsmath,amssymb,amsfonts}
\usepackage{algorithmic}
\usepackage{graphicx}
\usepackage{multirow}
\usepackage{subcaption}
\usepackage{booktabs}
\usepackage{textcomp}
\usepackage{xcolor}
\def\BibTeX{{\rm B\kern-.05em{\sc i\kern-.025em b}\kern-.08em
    T\kern-.1667em\lower.7ex\hbox{E}\kern-.125emX}}
\begin{document}

% \title{Getting Started with Scientific Writing in Overleaf}

\title{Assessing feasibility of few-shot learning in plant disease recognition}

\title{Image augmentation improves few-shot classification performance in plant disease recognition}

\author{\IEEEauthorblockN{Frank Xiao}
\IEEEauthorblockA{
%\textit{email}}
}}
\maketitle
\begin{abstract} 
    With the world population projected to near 10 billion by 2050, minimizing crop damage and guaranteeing food security has never been more important. Machine learning has been proposed as a solution to quickly and efficiently identify diseases in crops. Convolutional Neural Networks typically require large datasets of annotated data which are not available on demand. Collecting this data is a long and arduous process which involves manually picking, imaging, and annotating each individual leaf. I tackle the problem of plant image data scarcity by exploring the efficacy of various data augmentation techniques when used in conjunction with transfer learning. I evaluate the impact of various data augmentation techniques both individually and combined on the performance of a ResNet. I propose an augmentation scheme utilizing a sequence of different augmentations which consistently improves accuracy through many trials. Using only 10 total seed images, I demonstrate that my augmentation framework can increase model accuracy by upwards of 25\%.  
\end{abstract}

\section{Introduction }
﻿Plant disease poses large problem for food security. If not detected and treated, disease can spread quickly and destroy large areas of crops\cite{kranz1988measuring}. Additionally, not all diseases are widely known. Early detection allows for timely treatment, decreasing the chances for widespread crop failure. However, manual detection by humans can be a daunting task. With farms averaging over 1000 acres \cite{farmsize} of farmland, automation is becoming a necessity. While laboratory techniques have been employed to identify diseased plants, these approaches are very expensive and time consuming. Early detection, which requires scanning many plants, makes this approach infeasible. \par
%Statistics of plant disease affect on yield ^
﻿Machine Learning and Deep Learning have shown promise in the effort to quickly and efficiently categorize diseased crops. Taking advantage of high resolution cameras and advanced machinery, a program could quickly categorize large numbers of crops and flag those suspected of disease. \par

﻿One of the main issues facing the usage of machine learning and deep learning for disease recognition is the availability of data. Deep learning approaches such as convolutional neural networks typically require large amounts of labeled data to avoid overfitting. Due to the sheer variety of plants and diseases, obtaining a large enough dataset for each disease is infeasible, due to a large variety of issues, such as the weather, time constraints, or simply the sheer amount of work required to annotate so many samples. \par

﻿Image augmentation presents an elegant solution to this dilemma. Image augmentation schemes can create a richer and larger dataset from a limited set of seed images, dramatically reducing the number of images needed. This can allow higher performing machine learning models to be trained using less original data. \par

﻿Several studies have focused on few-shot learning in the agricultural field, but this is still an emerging field. Argüeso et al. used transfer learning in order to transfer knowledge from the source domain to the target domain, achieving over 90\% accuracy in the target domain\cite{argueso2020few}. Li et al. proposed a semi-supervised few-shot classification method based on transfer learning using mostly unlabelled samples\cite{li2021semi}. Chen et al. proposed local feature matching conditional neural adaptive processes based on meta learning that aimed to classify unknown categories with only a few annotated examples\cite{chen2021meta}. Hu et al. used a generative adversarial network to augment data in few-shot learning concerning tea leaves\cite{hu2019low}. \par

In this paper, I propose a few-shot classification method based on data augmentation techniques and transfer learning. This method uses a few labeled samples, along with knowledge transferred from the ImageNet dataset to the target domain. I apply this framework to a plant leaf disease detection task and demonstrate classification accuracy improvements of up to 25\%. 

%Key contributions(What you did):

% \begin{itemize}
%     \item 
% \end{itemize}

\section{Relevant Work} \label{review}
\subsection{Few-Shot Learning}

% Find the pivotal papers of the field (few-shot)
Few-shot learning refers to learning from small sets of data. Typically, a model is pre-trained on other data before being fine tuned on the target domain with only few data. Few-shot learning is very important as there are many scenarios where gathering the large datasets necessary for many models is simply not feasible. For this reason, few-shot learning has been gaining popularity in the recent years as people encounter new tasks which do not have the datasets required for some of the most popular models today.\par

Numerous strategies have been proposed to increase model accuracy in few-shot learning tasks. One popular method is data augmentation. By augmenting the amount of data available for fine tuning, model accuracy could be improved. Data augmentation in few-shot learning is a relatively new topic, but has been studied in several recent papers. For instance Antoniou et al. proposed a method to generate target sets using unlabeled data\cite{antoniou2019assume}. \par

There has also been significant work in using Generative Adversarial Networks (GANS) in order to generate data using the little data available. For example, Zhang et al. proposed a framework where GANs were used to generate new data in order to help classifiers sharpen the decision boundary\cite{zhang2018metagan}. This has also been applied specifically in the field of plant disease detection. Hu et al. use a conditional deep convolutional generative adversarial network in order to augment the dataset for few-shot tea leaf disease detection\cite{hu2019low}. Zhou et. al used a fine-grained GAN for local spot area data augmentation for few-shot grape leaf spot identification\cite{zhou2021grape}. \par

Transfer learning has also been a proposed method of increasing model accuracy in few-shot learning problems. Knowledge is transferred from a source domain to the target domain, where data is limited. Models are typically pretrained on the source domain before being fine-tuned on the few datapoints available in the target domain. Using this method takes advantage of similar problems with more readily available datasets. Work done in this field includes a method proposed by Sun et al. which learns to adapt a deep neural network for few-shot learning tasks \cite{sun2019meta}. Hu et al. proposed a transfer learning approach which builds on preprocessing feature vectors to more closely resemble a Gaussian-like distribution and leveraging this preprocessing using an optimal-transport inspired algorithm\cite{hu2021leveraging}  Yu et al. proposed a new transfer learning framework designed to fully utilize the auxiliary information from labeled base class data and unlabeled novel class data\cite{yu2020transmatch}. \par

%\subsection{Image augmentation in classification problems}
\subsection{Machine Learning in the Agricultural Sector}
Agriculture looks to be a prime field for influence by image recognition and machine learning in general. The visual nature of many problems naturally lends itself to an image recognition approach. Indeed, machine learning and image recognition in particular have been widely explored as a solution to many pressing problems in the field of agriculture.\par

For example, deep machine learning was used by Pound et. al in image-based plant phenotyping\cite{pound2017deep}. Similarly, Naik et. al developed a real-time phenotyping framework which used machine learning to determine plant stress severity in soybeans\cite{naik2017real}. Lee et. al designed a high-throughput plant phenotyping system using machine learning-based plant segmentation and image analysis\cite{lee2018automated}. Tsaftaris et. al highlight the importance of image analysis and processing when providing effective feature extraction for high throughput plant phenotyping\cite{tsaftaris2016machine}. \par

Plant breeding has also been a topic of interest of machine learning. Heslot et. al tested the predictive capabilities of gradient descent and machine learning models  on multiple datasets\cite{heslot2012genomic}. Parmley et. al use machine learning to develop a tool to assist with seed yield predictions as well as prescriptive cultivar development for targeted agro-management practices\cite{parmley2019machine}. Yoosefzadeh-Najafabadi et. al tested different common machine learning algorithms for predicting soybean seed yield using hyperspectral reflection\cite{yoosefzadeh2021application}.

One of the most researched intersections between agriculture and machine learning is the efficacy of image recognition models in detecting plant disease. Ferentinos used convolutional neural networks to identify diseased plant crops based on images of plant leaves\cite{ferentinos2018deep}. Too et. al demonstrated the efficacy of the DenseNet model for plant disease recognition\cite{too2019comparative}. Barbedo demonstrated the importance of large datasets for training convolutional neural networks to recognize plant disease\cite{barbedo2018impact}. Hughes et. al curated over 50,000 images of diseased and healthy plant leaves in order to enable computer vision approaches to help protect crops from disease\cite{hughes2015open}.

\section{Methodology } \label{methodology}

%Architecture diagram
\subsection{Dataset}
PlantVillage is a public dataset of over 50,000 images with 38 different classes of diseased and healthy plants leaves. Few-shot learning, unlike traditional deep learning approaches, only requires small amounts of data. For this experiment, 5 images were randomly sampled from a class and were used as the seed images. The other images were used for model validation.

%Diagram of the image augmentation schemes side by side@article
%Diagram of many images stemming from one source image
%An idea: Source image at the top of each column (5 columns), show 5-8 images that stem from each of the source images via stacked augmentations.

In this paper, only 10 images total were used to fine-tune the model on the target domain. In order to avoid overfitting, a data augmentation scheme was used. The data augmentation scheme used in this paper included many different transformations, all which had a certain chance of occurring each time the program was run. All the augmentations in the sequence were random in the sense that the augmentations had a set chance of occurring each time as well as that the augmentation would occur with differing strengths each run. \par

\subsection{Image Augmentation Schemes}
\begin{figure*}[!htb]
\centering
\includegraphics[scale=.3]{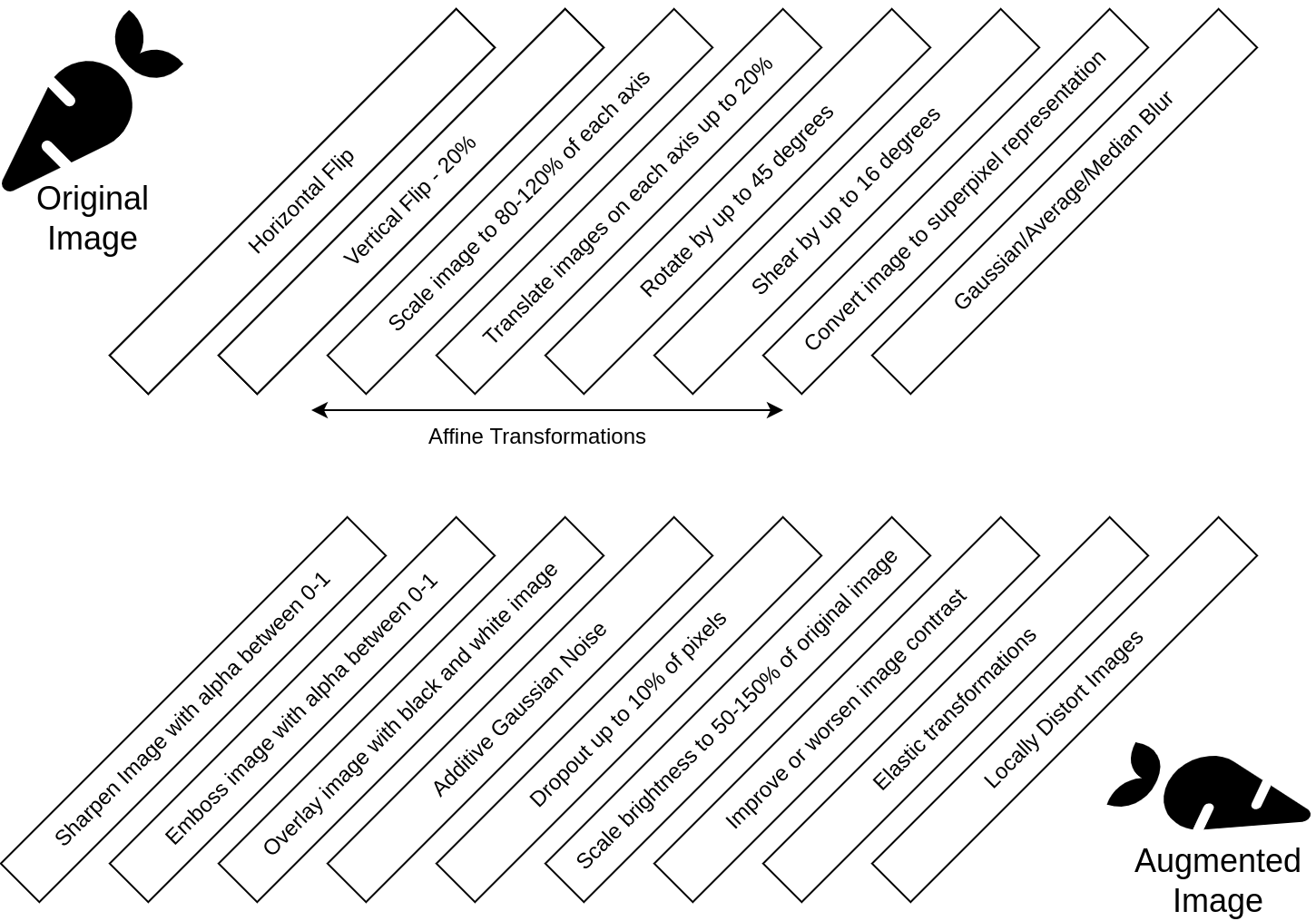}
\caption{Diagram of the augmentation scheme used}
\centering
\end{figure*}

The augmentation scheme used in this paper can be seen in Figure 1. All augmentations pictured had a 50 percent change of occurring unless otherwise specified.\par

The first augmentation that had a chance of occurring was a horizontal flip of the image. There was also a 20\% chance of a vertical flip afterwards. \par

The next augmentation that had a chance of happening were the affine transformations. Affine transformations were applied in a set (all or none) to images with 50\% chance. \par

The affine transformations consisted of several different operations. Image would be independently scaled from 80\% to 120\% of each of axis. The image would then be translated on either axis up to 20\% . Next, the image would be rotated up to 45 degrees in any direction and sheared up by to 16 degrees \par

After the affine transformations, there was a chance for an image to be replaced by its superpixel representation, varying from 20 to 200 superpixels for each image. There was also a chance for the image to be blurred by either Gaussian, average, or median blur. There was also a chance for the image to be either sharpened or embossed. \par

After this, a black and white image had a chance at being overlayed onto the original image. Additive Gaussian noise was sometimes applied to the image. There was also a chance of parts of the image being dropped out, as well as having the brightness and contrast adjusted. Finally, images could be locally distorted. \par

By running each of the 10 images through this sequence 500 times, 5010 total images were produced to be used in the fine tuning of the model. Since tests showed that further augmentation was largely ineffective at increasing the accuracy of the model on the target domain this quantity was deemed to be sufficient. Example images produced by the augmentation scheme can be seen in Figure 2. \par 

\begin{figure*}[!htb]
\centering
\includegraphics[scale=.5]{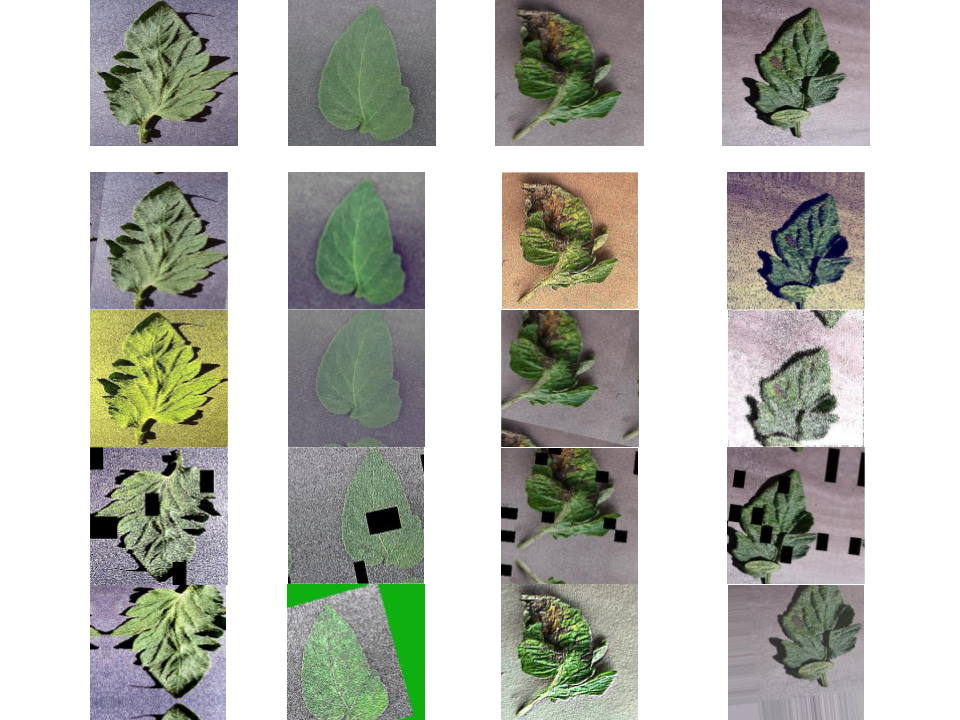}
\caption{Seed images and augmented images}
\centering
\end{figure*}

\begin{figure*}[!htb]
\centering
\includegraphics[scale=.5]{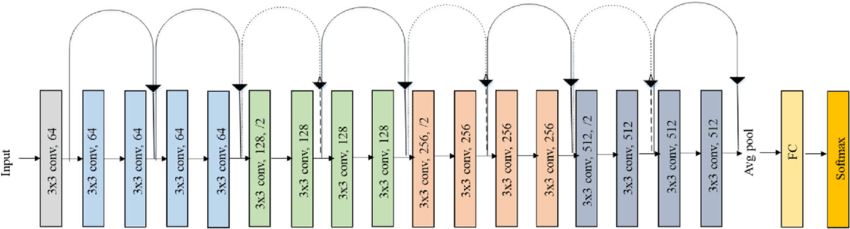} %make own?
\caption{Structure of a ResNet-18 \cite{resnet}}

\end{figure*}
\subsection{Classifier Architecture}

The model used in this experiment was a Residual Neural Network (ResNet), which has been widely used in image classification. The model architecture can be seen in Figure 1. As can be seen in the figure, the model has 4 layers, along with a classification layer added to the end. The Resnet-18 was chosen in order to combat overfitting due to the smaller nature of the dataset. The same model architecture was used for both the source and target domains.\par

\subsection{Training and Testing}
% Move training and testing stuff here
% include not as much information (mention gh link available)

In order to transfer knowledge from the source domain, in this case the ImageNet dataset, to the target domain, the model was first trained on the source domain. The Stochastic Gradient Descent (SGD) optimizer was used, as well as the categorical cross-entropy loss function. Due to the comparatively small number of images in the target domain, the convolutional base was then frozen and the classifier was fine tuned on the source domain. Project code can be found at https://github.com/frankyaoxiao/dataaug.

\section{Results} \label{results}
%Show your key findings. Make sure to state what the point of these findings are. Why are they useful? Don't just state what you did. MAKE NICE FIGURES
%add analysis of the figures
%these results illustrate how few datap -> high(ish) model acc
% abalation study -> analyze how much everything is contributing 

\begin{table*}[h!]
\centering
\begin{tabular}{cccccc}
\toprule
Plant leaf type & Baseline & Augmented Trial 1 & Augmented Trial 2 & Augmented Trial 3 & Trial Average \\ \midrule
Apples & 72.4\% & 84.3\% & 82.5\% & 84.1\% & 84.0\% \\
Cherry & 55.1\% & 83.3\% & 91.7\% & 90.4\% & 88.4\% \\
Pepper & 56.2\% & 84.3\% & 83.7\% & 83.5\% & 83.8\% \\
Potatoes & 86.8\% & 89.4\% & 96.7\% & 95.8\% & 94.0\% \\
Strawberry & 70.8\% & 98.0\% & 98.2\% & 98.4\% & 98.2\% \\
Tomato & 70.7\% & 84.4\% & 79.9\% & 85.2\% & 83.2\% \\
\bottomrule
\end{tabular}
\caption{Accuracy of model before and after augmentations were applied} 
\label{table:1} 
\end{table*}

Throughout the experiments, a clear improvement can be seen from the use of the proposed augmentation schemes. Table 1 presents the overall model performance for each plant species. The baseline column depicts the model accuracy when trained only on the seed images. The Augmented Trials 1-3 columns show the accuracy of the model after being trained on the augmented dataset. It can be observed that the augmentations increased the overall accuracy of the model for every class. \par

It can also be noted that the percentage increase in accuracy varied from class to class. The augmentation scheme improved performance by upwards of 25\% when used on seed images of cherry leaves, while only improving performance by around 10\% when used on seed images of apple leaves. Accuracy fluctuations between different trials can probably be attributed to the different seed images chosen each round. \par

\begin{table*}[h!]
\centering
\begin{tabular}{ccc}
\toprule
Augmentation type & Accuracy & Percentage Change \\
\midrule
None              & 70.4\% &  0\% \\
Rotation          & 81.4\%  & +11.0\% \\
Mirror + Flip     & 74.3\%  & +3.9\% \\
Channel Shift     & 62\%  & -8.4\%   \\
Gaussian Noise    & 72.4\% & +2.0\% \\
Blur              & 74\%  &  +3.6\% \\
\bottomrule
\end{tabular}
\caption{Accuracy of model after individual augmentations on a set of 5 diseased and 5 healthy apple leaves}
\label{table:2}
\end{table*}

To examine the individual effect of different augmentations, I tested model performance before and after using only a single augmentation. Table 2 shows the model accuracy after only one augmentation was performed on 10 seed images of apple leaves. In most cases, individual augmentations yielded marginal improvements, with the noticeable exceptions in the rotation augmentation and the channel shift augmentation. The rotation augmentation yielded over a 10\% increase in model accuracy, while the channel shift augmentation actually decreased accuracy. This makes sense intuitively because the primary difference between a healthy and an unhealthy leaf is the color of the leaf, which is what channel shifting changes. \par

\begin{table*}[h]
\centering
\begin{tabular}{ccc}\hline
\toprule
Plant type & No Pretraining & With Pretraining \\
\midrule
Apple      & 67.5\%         & 73.8\%           \\
Cherry     & 53.2\%         & 55.2\%           \\
Strawberry & 58.5\%         & 69.4\%   
\\
\bottomrule
\end{tabular}
\caption{Averaged accuracy of ResNet trained on 10 seed images with and without transfer learning}
\label{table:3}
\end{table*}

Table 3 shows the effects of pretraining the model on the ImageNet database. In all trials, the ResNet was trained on 10 seed images of the specified plant type, with the only difference being whether knowledge had been transferred from ImageNet. It can be seen that transfer learning increased model accuracy in every case, in the case of strawberry leaves upwards of 10\%. \par

Through the experiments undertaken in this paper, it can be seen that usage of an image augmentation scheme demonstrates noticeable improvements in plant disease few shot learning. In can be shown that most individual augmentations increased the accuracy of the model, with the notable exception of color shifting augmentations. It was also seen that pretraining the model on ImageNet yielded slight accuracy improvements. This demonstrates the viability of few shot learning in plant disease detection.\par 

% \section{Key Contributions} \label{contributions}
% This is an optional section which concisely states the key research contributions of the paper (bullet points often work well here)

\section{Future Research Directions} \label{results}
%This is sometimes formulated as `limitations'. In essence, this should sum up anything that wasn't successful or that you didn't get to in your project and should frame it as opportunities for future research. You should also mention other research projects that your work can enable. 
There are many directions future research can take in this field. More tests could be done on different types of plants to confirm the accuracy increase that results from transfer learning. Different convolutional neural network architectures could be tested to determine which type would yield the maximum accuracy for few shot learning in plant disease recognition. Different variations of augmentation sequences could be tested to discover increasingly accurate augmentation schemes.\par

Future viable research routes also include leveraging a GAN in order to further generate data in a few shot scenario after being trained on already augmented seed images. Conditional GANs could also be trained on a general dataset and then be conditioned on the few data to generate a particular species of leaves.\par

\section{Conclusion }
Plant disease poses a large problem for food security. Machine learning and image recognition in particular have been identified as solutions to quickly and efficiently discover diseased crops. This paper proposed an few shot classification method based on data augmentation and transfer learning. Using my data augmentation framework, I showed improvements of upwards of 25\% when using a residual neural network trained with only 10 seed images. The significant improvements in accuracy afforded by this novel few-shot learning approach constitute a noteworthy stride towards practical deep learning tools for plant disease recognition. 

\bibliographystyle{IEEEtran}
\bibliography{bibliography}

\end{document}